\documentclass{article}

\usepackage{amssymb}
\usepackage[linesnumbered,vlined,ruled]{algorithm2e}
\usepackage{amsmath}
\usepackage{url}
\usepackage{graphicx}
\usepackage{multirow}
\usepackage{rotating}
\usepackage{tabularx}

\usepackage[french]{babel} 
\usepackage[utf8]{inputenc} 
\usepackage{lmodern} 
\usepackage{xspace} 
\AddThinSpaceBeforeFootnotes 
\FrenchFootnotes 

\title{Planification en temps réel \\ avec agenda de buts et sauts}
\author{Damien Pellier, Bruno Bouzy, Marc M\'etivier\\
  Laboratoire d'Informatique de Paris Descartes \\
  45, rue des Saints Pères, 75270 Paris\\
  \texttt{\small \{damien.pellier, bruno.bouzy, marc.metivier\}@parisdescartes.fr}
}
\date{Juin 2011}

\begin{document}

\maketitle

\begin{abstract}
Cet article propose deux contributions pour améliorer la phase de sélection d'actions dans le cadre de la planification temps réel. Tout d'abord, la première amélioration s'appuie sur un agenda de buts pour classer les buts par ordre de difficulté croissante et tente ensuite de les résoudre de manière incrémentale. La seconde amélioration consiste à effectuer des sauts, c'est-à-dire à choisir à chaque pas de décision une séquence d'actions qui sera exécutée et non plus une seule action. Pour évaluer ces deux améliorations, nous avons développé un algorithme de planification temps réel dans lequel la phase de sélection d'actions peut être guidée par un agenda de buts et également capable d'exécuter un saut à chaque pas de décision. Les résultats expérimentaux, réalisés sur les problèmes de planification classiques issus des différentes compétitions de planification, montrent que l'agenda de buts ainsi que les sauts améliorent de manière importante la sélection d'actions dans le contexte de la planification temps réel. Utilisées simultanément, ces deux améliorations permettent d'augmenter de manière significative la qualité des plans trouvés ainsi que la vitesse pour les obtenir.
\end{abstract}

\section{Introduction}

Le domaine de la planification peut être découpé en deux grandes familles d'algorithmes que l'on qualifie de <<~{\it off-line}~>> et de <<~{\it on-line}~>>. Les algorithmes {\it off-line} cherchent dans un premier temps un plan solution, puis dans un second à exécuter le plan trouvé. Il est important de noter que de nombreuses heuristiques performantes, admissibles ou pas, ont été développées  pour guider la recherche efficacement vers le but pour ce type d'approche. Ces fonctions heuristiques sont calculées pour la plupart en s'appuyant sur une version relaxée d'une structure de donnée appelée graphe de planification \cite{blum:97}. La recherche proprement dite s'appuie, elle, sur différentes variantes de recherche de type informé, e.g., Enforced Hill Climbing \cite{hoffmann:01}, Weighted-{\it A*} \cite{pohl:70}. Cette première famille d'algorithmes trouve généralement des plans solutions très rapidement même s'ils ne sont pas toujours optimaux. Parmi l'ensemble des techniques algorithmiques développés, une technique particulièrement efficace est la planification dirigée par un agenda de buts \cite{koehler:00}. Cette technique effectue une recherche incrémentale en résolvant à chaque incrément un sous ensemble des propositions du buts. Cette approche qui applique le principe de \emph{diviser pour mieux régner} a été implantée dans ne nombreux planificateurs permettant ainsi d'augmenter de manière significative leur performance \cite{hoffmann:01,hoffmann:04,richter:08}.

Inversement, les planificateurs {\it on-line} n'exécutent pas un plan solution mais choisissent seulement la ou les meilleurs actions à exécuter en temps constant jusqu'à ce que le but soit atteint. La littérature distingue deux approches: la première fondée sur les processus décisionnels de Makov (MDP) et appliquée aux problèmes non-déterministes, e.g., \cite{barto:95,hansen:01} et la seconde s'appuyant sur les algorithmes de recherche temps réel ({\it RTS}), e.g., \cite{korf:90}. Si la première approche a été récemment appliquée à la planification \cite{fabiani:07}, la seconde a toujours été fortement associée au développement des jeux vidéo et au problème de {\it path-finding}. De nombreux algorithmes ont été développés dans se cadre, e.g., mini-min \cite{korf:90}, {\it B-LRTA*} \cite{bonet:97}, $\gamma$-Trap et {\it LRTS} \cite{bulitko:06}, {\it LRTA*$_{LS}$} \cite{hernandez:09}, ou encore {\it LSS-LRTA*} \cite{koenig:09}. Dans ces approches, meilleure est la sélection d'actions, meilleure est la qualité du plan solution trouvé. Étant donné que le temps alloué pour choisir l'action à exécuter est limité, ces algorithmes explorent un sous-ensemble de l'espace de recherche autour de l'état courant. Par conséquent, le plan solution finalement trouvé et exécuté n'est pas nécessairement optimal et le planificateur peut échouer dans sa recherche d'un plan solution, e.g., en exécutant une action qui le conduit dans un puits de l'espace de recherche. Ces observations montre qu'il existe une large place à l'amélioration du mécanisme de sélection d'actions notamment en l'adaptant et en le généralisant à la planification. Pour y parvenir, nous proposons d'utiliser une technique d'agenda de buts afin de guider le processus de sélection d'actions. Pour autant que nous le sachions, cette technique n'a pas encore été utilisée dans ce contexte. Dans cet article, nous nous intéressons à deux améliorations du processus de sélection dans le cadre de la planification temps réel. La première amélioration, notée {\it I} pour planification incrémentale, s'appuie sur un agenda de buts pour classer les buts par ordre de difficulté croissante et les résoudre de manière incrémentale. La seconde amélioration, notée {\it J}, consiste à choisir à chaque pas de décision une séquence d'actions qui sera exécutée et non plus une seule. On parle alors de saut. Pour évaluer ces deux améliorations, nous avons développé un algorithme de planification temps réel, appelé {\it LRTP} ({\it Learning Real-Time Planning}) fondé sur le modèle de {\it LRTS} \cite{bulitko:06}. Les sauts ({\it J}) et la planification incrémentale ({\it I}) peuvent étre activés ou pas pour améliorer le processus de sélection d'actions.

Cet article est organisé de la manière suivante: la section 2 donne une vue d'ensemble des travaux dans le domaine des algorithmes de type {\it RTS}; la section 3 présente les principales techniques d'agenda de buts développées et la section 4 introduit les notations utilisées et le pseudo-code de {\it LRTP} avec les deux améliorations {\it I} et {\it J}. La section 5 présente des résultats expérimentaux montrant que l'agenda de buts ainsi que les sauts permettent d'améliorer de manière significative le processus de sélection d'actions de la planification temps réel. Finalement, avant de conclure, la section 6 discute les résultats obtenus.

\section{Recherche temps réel} \label{sec:real-time-search}

Par opposition aux algorithmes classiques de recherche {\it off-line} qui recherchent un plan solution avant de l'exécuter, les algorithmes {\it RTS} entrelacent planification et exécution. Ces algorithmes ont été particulièrement étudiés dans le contexte de ce que la littérature appelle la recherche centrée agent. Le principe est le suivant: un agent exécute à plusieurs reprises une tâche; à chaque fois que la tâche est exécutée, on parle d'épisode; à tout moment au cours d'un épisode, l'agent est dans un état, appelé état courant, et effectue une recherche locale à partir de cet état afin de choisir la meilleure action pour atteindre le but qui lui a été confié. Les états construits au cours de la phase de recherche locale sont appelés les états explorés. Il est important de noter que toutes les recherches locales sont réalisées en temps borné ce qui confère la propriété temps réel à ces algorithmes. Après avoir sélectionné une action, l'agent l'exécute modifiant ainsi son état courant. L'ensemble des états dans lesquels s'est trouvé l'agent sont appelés les états visités. Un épisode se termine lorsque l'agent se trouve dans un état satisfaisant le but ou lorsque l'agent se trouve dans un état puits (l'agent a échoué). Finalement, l'agent relance un nouvel épisode et ce tant qu'un plan solution optimal n'est pas trouvé.

Les algorithmes de recherche temps réel peuvent être considérés avec ou sans apprentissage. Sans apprentissage, l'efficacité des algorithmes {\it RTS} dépendent uniquement de la qualité de la recherche locale à choisir une bonne d'action. En revanche, avec apprentissage, les algorithmes {\it RTS} mettent à jour la valeur de la fonction heuristique associé aux états visités lorsque certaines conditions sont vérifiées et la qualité du plan trouvé augmente au fur et à mesure des répétitions.

L'article fondateur des algorithmes de type {\it RTS} est \cite{korf:90} intitulé {\it Real-Time Heuristic Search}. De nombreux algorithmes ont été développés à partir de ces travaux. Nous pouvons citer notamment: Real-Time {\it A*} ({\it RTA*}) qui construit un arbre comme {\it A*} mais en temps constant; la version avec apprentissage de ce même algorithme qui est appelé {\it LRTA*} pour {\it Learning Real-Time A*}; des travaux plus récents qui ont cherché en particulier à améliorer les performances en se concentrant sur l'apprentissage, e.g., FALCONS \cite{furcy:00}, {\it LSS-LRTA*} \cite{koenig:09}; ou encore $\gamma$-Trap and {\it LRTS} \cite{bulitko:06} qui proposent un modèle unifié pour l'apprentissage dans le contexte de la recherche temps réel.

\section{Planification incrémentale}
\label{sec:agenda-driven-planning}

Une des manières de guider la recherche d'un plan solution consiste à définir l'ordre dans lequel les propositions atomiques ou sous-buts constituants le but global doivent être atteints. Cette technique, appelée planification incrémentale ou dirigée par un agenda de buts, améliore les performances de la recherche en guidant le planificateur vers la construction progressive d'un plan solution. Un ordre idéal est un ordre dans lequel chaque proposition atomique ne peut pas être invalidée pas une proposition précédemment vérifiée. Autrement dit, si aucune des propositions atomiques n'invalide une proposition ordonnée précédemment dans l'agenda de buts au cours de la recherche d'un plan solution, cela signifie que l'agenda a découpé le but global en sous-buts pouvant être résolus incrémentallement de manière indépendante par ajouts successifs. Inversement, si plusieurs sous-buts sont invalidés après avoir été vérifiés, l'ordre fournit par l'agenda de buts n'est pas informatif car les efforts consentis précédemment pour les atteindre s'avèrent être inutiles. Par conséquent, l'efficacité de tout algorithme de planification incrémentale dépend fortement de la qualité de l'agenda qui guide la recherche d'un plan solution.

Le problème de la construction d'un agenda de but peut se résumer de la manière suivante: étant donné une conjonction de propositions atomiques représentant le but à atteindre, est il possible de détecter une relations d'ordre entre les sous-ensembles de propositions atomiques du but ? Un des premiers à s'être penché sur cette question est Koehler \cite{koehler:00}. Il a introduit la notion {\it d'ordre raisonnable} qui définit que deux sous-buts $A$ et $B$ sont raisonnablement ordonnés s'il n'est pas possible d'atteindre un état tel que $A$ et $B$ soient vérifiés, à partir d'un état où seulement $B$ est vérifié, sans avoir temporairement invalidé $B$. Dans une telle situation, $A$ doit être raisonnablement ordonné avant $B$ pour limiter des recherches inutiles.
Trois principales extensions de ce travail préliminaire ont été proposées. La première, appelé {\it landmarks planning}, a été introduite par \cite{hoffmann:04}. Cette extension n'ordonne pas seulement le but global, mais également les sous-buts qui apparaissent nécessairement au cours de la recherche d'un plan solution. Les sous-buts sont appelés des {\it landmarks}. La principale caractéristique des {\it landmarks} est qu'ils doivent être vérifiés à un instant donné quel que soit le chemin solution emprunté. Cette approche a été implémentée avec succès dans le planificateur LAMA \cite{richter:08}.

La seconde extension a été présentée par Hsu \cite{hsu:05}. La construction de l'agenda de buts peut se résumer en trois étapes: (1) construire un graphe de planification $G$ \cite{blum:97} à partir de l'état initial jusqu'à ce que le point fixe du graphe soit atteint sans calculer les exclusions mutuelles ; (2) extraire du graphe $G$ un plan relaxé en ignorant les effets négatifs des actions de manière identique au calcul de la fonction heuristique réalisé par le planification FF \cite{hoffmann:01} ; (3) déterminer un ordre entre chaque paire $A$ et $B$ de sous-buts en examinant toutes les actions de $G$ qui rendent $B$ vraie. Il est important de noter que toutes les relations d'ordre partiel détectées par l'approche de Koehler \cite{koehler:00} le sont également par cette approche. Cette approche est donc strictement meilleure que la précédente.

Finalement, Botea \cite{botea:09a} a récemment proposé un algorithme de planification incrémentale guidé par un agenda de buts dans le cadre de la planification temporelle, avec buts étendus et prise en compte d'évènements externes.

\section{LRTP avec agenda de buts et sauts}
\label{sec:lrtp}

Cette section introduit les notations et l'implémentation de {\it LRTP} ({\it Learning Real-Time Planning}).

\subsection{Notations}

Nous nous plaçons dans le cadre de la planification temps réel au sens STRIPS  \cite{finke:71}. Tous les ensembles considérés sont supposés finis. Un {\it état} $s$ est défini par un ensemble de propositions logiques. Une {\it action} $a$ est représentée par un tuple $a = (pre(a), add(a), del(a))$ où $pre(a)$ définit les préconditions de l'action et $add(a)$ et $del(a)$ respectivement ses {\it effets} positifs et négatifs. Chacun des ensembles $pre(a)$, $add(a)$ et $del(a)$ sont des ensembles de propositions. Le résultat de l'application d'une action $a$, notée comme une séquence d'une seule action, dans un état $s$ est modélisé par la fonction de transition $\gamma$ comme suit:
\begin{equation*}
  \gamma(s,\langle a \rangle) =
  \begin{cases}
    (s - del(a)) \cup add(a)  & \text{si} \ pre(a) \subseteq s\\
    \text{indéfini} & \text{sinon}
  \end{cases}
\end{equation*}
Le résultat de l'application d'un plan $\pi$, i.e., une séquence de plus d'une action $\pi = \langle a_1, \ldots, a_n \rangle$, à un état $s$ se définit récursivement par la fonction $\gamma(s,  \langle a_1, \ldots, a_n \rangle) = \gamma(\gamma(s, \langle a_1, \ldots, a_{n-1} \rangle), \langle a_n \rangle)$. Notons que l'application d'un plan vide n'a aucune conséquence, i.e., $\gamma(s, \langle \rangle) = s$. Un {\it problème de planification} $(A, s_0, g)$ est un tuple où $A$ est un ensemble d'actions, $s_0$ l'état initial et $g$ le but. Un plan  $\pi = \langle a_1, \ldots, a_n \rangle$ avec $a_1, \ldots, a_n \in A$ est solution d'un problème de planification $(A, s_0, g)$ si $g \subseteq \gamma(s_0, \pi)$. Le temps alloué à la sélection d'actions est borné par une constante $t_d$, appelé {\it temps de décision}. Finalement, $t_g$ est le temps global alloué à la recherche.

Dans la suite de cet article, nous nous focalisons uniquement sur la phase de sélection d'actions. Par conséquent, nous ne traitons pas des aspects liés à l'apprentissage. Le mécanisme d'apprentissage utilisé dans {\it LRTP} est identique à celui de \cite{bulitko:06}.

\subsection{Algorithme}

Le pseudo-code de {\it LRTP} est décrit par l'algorithme~\ref{Algo:LRTP}. Il prend en entrée un problème de planification  $(A, s_0, g)$, le temps de décision $t_d$ ainsi que le temps global alloué à la recherche $t_g$. Dans un premier temps, {\it LRTP} exécute une boucle (ligne 1) dans laquelle il exécute itérativement des épisodes tant que le temps global alloué $t_g$ n'est pas dépassé. $s_0$ représente l'état initial à partir duquel débute les différents épisodes. $s$ représente l'état courant à partir duquel la phase de planification est initiée, et $s_r$ est l'état courant du monde à partir duquel les actions sont exécutées (ligne 2). Dans un second temps, {\it LRTP} entre dans la boucle de décision (ligne 3). À chaque itération, {\it LRTP} cherche la meilleure séquence d'actions $\pi' = \langle a_{1}, \ldots, a_{n} \rangle$ qui le conduit au but $g$. La recherche est réalisée par la procédure {\it IASA*} avec l'amélioration {\it I} (ligne 4) et par la procédure {\it ASA*} (ligne 5) dans le cas contraire.

Lorsque l'amélioration {\it J} est active (ligne 7), {\it LRTP} concatène la séquence d'actions $\pi'$ au plan courant $\pi$ et met à jour son état courant $s$ avec la fonction $\gamma(s, \pi')$. Avec cette extension, {\it LRTP} saute de son état courant à l'état résultant de l'application de la dernière action de $\pi'$ qui devient alors le nouvel état courant $s$. Lorsque l'amélioration {\it J} est inactive (ligne 8), {\it LRTP} concatène seulement la première action $a_1$ de $\pi'$ au plan courant $\pi$ de l'épisode et met à jour $s$ en utilisant la fonction $\gamma(s, \langle a_1 \rangle)$. Par la suite, quelque soit l'état d'activation de l'amélioration {\it J}, {\it LRTP} dépile et supprime la première action du plan $\pi$ et l'exécute. Finalement, $s_r$ est mis à jour (ligne 10).

Lorsque l'amélioration {\it J} est active, {\it LRTP} ne peut pas modifier les actions qui sont stockées dans son tampon. Ceci a pour résultat que $s_r$ et $s$ peuvent être différents. {\it LRTP} tire parti du fait que plusieurs actions aient été mises en zone tampon en une seule décision pour accumuler un crédit de temps qu'il utilisera pour sa prochaine décision. Cette technique permet à {\it LRTP} d'effectuer une recherche plus approfondie de l'espace de recherche local et d'augmenter ainsi la qualité des séquences d'actions obtenues.

\begin{algorithm}[!t]
\DontPrintSemicolon
\caption{LRTP($A,s_0,g, t_d, t_g$)}
\label{Algo:LRTP}
\While{$elapsed\_time < t_g$}{
  $\pi \leftarrow \langle \rangle$, $s \leftarrow s_0$, $s_r \leftarrow s_0$ \;
  \While{$elapsed\_time < t_g$ and $g \not\subseteq s_r$}{
    \lIf{I is on}{
      $\pi' \leftarrow $ IASA*($A, s, g, t_d$)\;
    }
    \lElse{
      $\pi' \leftarrow $ ASA*($A, s, g, t_d$)\;
    }
    {
      Let $\pi' = \langle a_{1}, \ldots, a_{k} \rangle$ \;
      \lIf{J is on}{
        $\pi \leftarrow \pi \cdot \pi'$, $s \leftarrow \gamma(s, \pi')$ \;
      }
      \lElse{
        $\pi \leftarrow \pi \cdot a_{1}$, $s \leftarrow \gamma(s, \langle a_1 \rangle)$ \;
      }
      Execute and remove the first action $a$ of $\pi$ \;
      $s_r \leftarrow \gamma(s_r, \langle a \rangle)$ \;
    }
  }
}
\end{algorithm}

\subsection{Sélection d'actions}

L'objectif de l'étape de la sélection d'actions est de calculer en temps limité la séquence d'actions la plus proche du plan solution optimal pour le problème de planification donné. Pour atteindre cet objectif, nous nous appuyons sur l'algorithme {\it A*} \cite{hart:68}. Le pseudo-code décrivant cette procédure, appelée {\it ASA*}, est décrite par l'algorithme~\ref{Algo:ASA*}. Cet algorithme de sélection d'actions présente la procédure de sélection non guidée par un agenda de buts. Son principe est simple. Lorsque le temps de décision $t_d$ est écoulé, {\it ASA*} choisit l'état le plus proche du but parmi la liste des états contenus dans la liste $S_{open}$ (ligne 1). La séquence actions retournée est le chemin reliant l'état initial $s$ à l'état feuille choisi. La sélection de l'état le plus proche du but dans $S_{open}$ s'effectue en trois étapes. Premièrement, {\it ASA*} sélectionne les états ayant la valeur de $f$ la plus basse (ligne 2). Ces états sont stockés dans $S_f$. Puis dans un second temps, {\it ASA*} choisit parmi les états de $S_f$ ceux qui possèdent la plus petite valeur de $g$ et les stocke dans $S_g$ (ligne 3). Finalement, {\it ASA*} désigne de manière aléatoire le nouvel état courant $s$ parmi les états $S_g$ (ligne 4). L'idée sous-jacente ici est de donner la priorité aux états les plus proches du but mais également aux états les plus proches de l'état initial afin d'augmenter la robustesse des séquences d'actions retournées. Cette technique n'est pas nouvelle \cite{bulitko:06}.

\begin{algorithm}[!h]
\DontPrintSemicolon
\caption{ASA*($A,s_0,g,t_d$)}
\label{Algo:ASA*}
$S_{open} \leftarrow$ A*($A, s_i, g_i, t_d)$\;
Let $S_{f}$ the set of states with the lowest $f$-value of $S_{open}$ \;
Let $S_{g}$ the set of states with the lowest $g$-value of $S_{f}$ \;
Choose randomly a state $s$ from $S_g$ \;
\Return the plan from $s_0$ to $s$ \;
\end{algorithm}

Le pseudo-code {\it IASA*} décrit la procédure de sélection d'actions dirigée par les buts (cf. Algorithme~\ref{Algo:IASA*}). Dans un premier temps, la procédure de sélection calcule l'agenda de buts en s'appuyant sur les travaux de Hsu \cite{hsu:05}, i.e., en déterminant toutes les relations d'ordre entre les propositions atomiques du buts (ligne 2). Le but $g$ est dorénavant une séquence de propositions atomiques $\langle g_1, \ldots, g_n \rangle$ telle que $g = \bigcup_{i=1}^{n} g_i$. Puis, pour chaque proposition atomique $g_j$ et ce tant que le temps alloué à la recherche locale n'est pas dépassé, la procédure de recherche locale {\it ASA*} est lancée pour déterminer un plan ayant pour état initial $s_i$ (initialement assigné à $s_0$) et permettant d'atteindre incrémentalement le but  $g_i = \bigcup_{j=1}^{i} g_j$ (lignes 4-5). Le plan solution pour le sous-but $g_i$ est alors ajouté à la séquence d'actions en cours de construction et l'état $s_i$ est mis à jour en appliquant la fonction $\gamma(s_i, \pi')$ (ligne 7). Finalement, si toutes les propositions atomiques constituant le but sont vérifiées ou si le temps de décision alloué est dépassé, la recherche prend fin et la procédure retourne la séquence d'actions calculée.

\begin{algorithm}[!h]
\DontPrintSemicolon
\caption{IASA*($A,s_0,g,t_d$)}
\label{Algo:IASA*}
$\pi \leftarrow \langle \rangle$, $i \leftarrow 1$, $s_i \leftarrow s_0$ \;
$\langle g_1, \ldots, g_n \rangle \leftarrow$ RelaxedPlanOrdering($A, s_0, g$) \;
\While{$elapsed\_time < t_d$ and $i \leq n$}{
  $g_i \leftarrow \bigcup_{j=1}^{i} g_j$ \;
  $\pi' \leftarrow $ ASA*($A, s_i, g_i, t_d - elapsed_time)$\;
  $\pi \leftarrow \pi \cdot \pi'$ \;
  $s_i \leftarrow \gamma(s_i, \pi')$ \;
  $i \leftarrow i + 1$ \;
}
\Return $\pi$ \;
\end{algorithm}

\section{Expériences}
\label{sec:experiments}

L'objectif de ces expériences est d'évaluer les performances de {\it LRTP}, avec ou sans l'amélioration {\it I}, et avec ou sans l'amélioration {\it J}, dans des problèmes de planification classique tirés des Compétitions Internationales de Planification (IPC). Nous utilisons la fonction heuristique non-adminissible de FF \cite{hoffmann:01} pour conduire la recherche. L'utilisation d'une heuristique non-admissible pour la recherche en temps réel n'est pas nouveau \cite{Shimbo:03}. Nous avons quatre algorithmes à évaluer : {\it LRTP} sans  amélioration, {\it} LRTP avec la planification incrémentale ({\it LRTP+I}), {\it  LRTP} avec sauts ({\it LRTP+J}), et {\it LRTP} avec les deux améliorations ({\it LRTP+IJ}).

\subsection{Description, paramètres et mesures de performances}

Nos expériences consistent à exécuter chaque algorithme sur un problème spécifique, avec un temps de décision défini et pour un nombre défini d'épisodes. Pour chaque algorithme, une expérience fournit les performances suivantes: (i) le pourcentage de réussite, i.e., le pourcentage d'épisodes dans lesquels l'objectif a été atteint et (ii) la longueur moyenne des plans trouvés.

Dans un premier temps, les quatre algorithmes sont évalués sur un problème spécifique issu d'IPC-5, le problème 15 du domaine Rovers, choisi pour mettre en évidence la façon dont les différents algorithmes se comportent selon le temps de décision. Chaque expérience comprend 10 épisodes. Tout épisode se termine si l'algorithme atteint l'objectif ou a exécuté 400 actions.

Dans un deuxième temps, les algorithmes sont évalués sur tous les problèmes du domaine Rovers avec des temps de décision différents. Chaque expérience  contient  100  épisodes. Les épisodes se terminent si l'algorithme atteint  l'objectif ou si 500 actions  ont été exécutées. Les résultats fournis  sont  les  performances  moyennes  sur chaque problème du domaine.

Enfin, dans un troisième temps, nous présentons les résultats obtenus par les quatre algorithmes sur tous les problèmes de 15 domaines extraits des différentes compétitions de planification. Un temps de décision spécifique est utilisé pour chaque domaine. Comme précédemment chaque expérience contient 100 épisodes et chaque épisode se termine si l'algorithme atteint l'objectif ou si 500 actions ont été exécutées. Les résultats fournis sont les performances moyennes sur chaque domaine. Dans ces résultats, si un algorithme n'a jamais atteint l'objectif en un problème, la longueur de son plan est fixé à 500.

Tous les tests ont été effectués sur un processeur Intel Core 2 Quad 6600 (2,4 GHz) avec 4 Go de RAM.

\subsection{Rovers 15}

Ce paragraphe présente les résultats obtenus par les quatre algorithmes pour le quinzième problème du domaine Rovers, Rovers 15. Il montre le pourcentage de réussite et la qualité de la sélection d'actions.

\subsubsection{Pourcentage de succès}

\begin{figure}
\begin{center}
\includegraphics[scale=0.70]{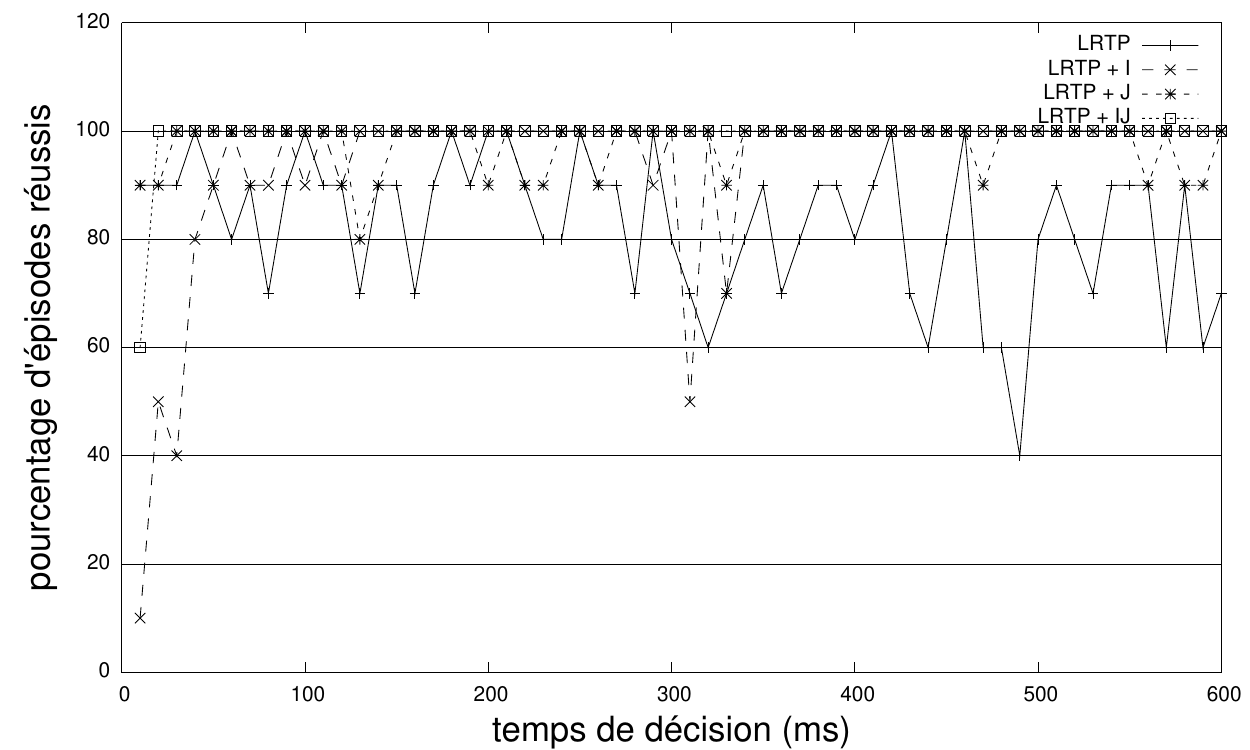}
\caption{Pourcentage de réussite selon le temps de décision (pb. 15 de Rovers, IPC5).}
\label{Fig:success-rover15}
\end{center}
\end{figure}

La figure~\ref{Fig:success-rover15} montre le pourcentage de réussite de chaque algorithme pour le problème 15 du domaine Rovers en fonction du temps de décision. Le pourcentage de réussite mesure le pourcentage d'épisodes dans lesquels l'algorithme trouve un plan de solution. {\it LRTP} sans amélioration obtient 100\% de réussite pour certains temps de décision spécifiques. Toutefois, il ne montre pas un comportement stable en fonction du temps de décision. Le plus souvent, son pourcentage de réussite reste entre 70\% et 90\%. {\it LRTP+I} obtient 100\% de réussite pour tout temps de décision supérieur à 130 ms, à l'exception de trois temps de décision spécifiques : 290 ms, 310 ms  et  330 ms. {\it LRTP+J} obtient 100\% de réussite avec plus de 30 ms de temps de décision, mais tandis que le temps de décision augmente, une  légère baisse des performances se produit parfois. Les  meilleures performances sont obtenues en utilisant les deux améliorations ensemble : un pourcentage de réussite de 100\% est obtenu pour des temps de décision supérieurs à 20 ms.

\subsubsection{Qualité de la sélection d'actions}

\begin{figure}
\begin{center}
\includegraphics[scale=0.70]{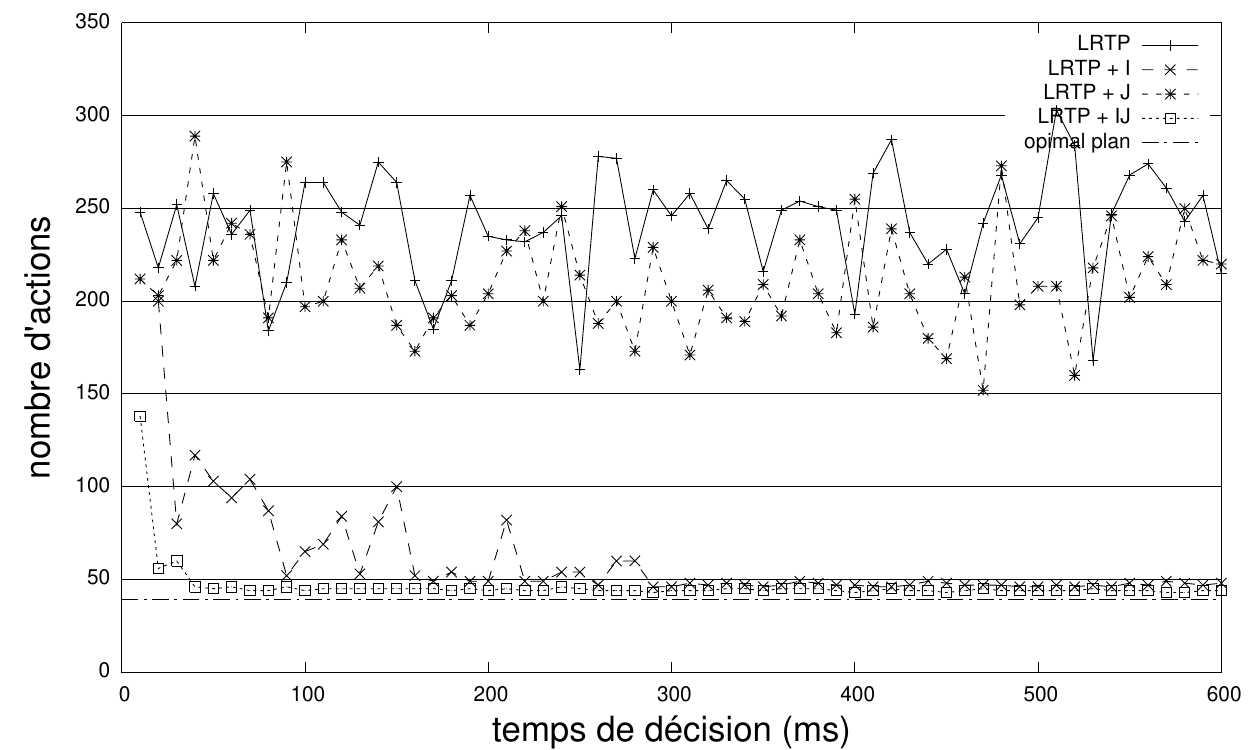}
\caption{Longueur du premier plan en fonction du temps de décision (pb. 15 Rovers IPC5).}
\label{Fig:1stPlanLength}
\end{center}
\end{figure}

La figure~\ref{Fig:1stPlanLength} montre la longueur du premier plan solution obtenu par chacun des algorithmes. Nous observons trois plages de temps de décision. Pour les temps de décision très petits, le processus de sélection d'actions sélectionne une action loin d'être optimale, et la longueur du premier plan solution est élevée. Pour les temps de décision moyens, le processus de sélection d'actions sélectionne une action à chaque pas de temps résultant en des plans solutions plus courts. Pour les temps de décision suffisamment longs, les plans solutions minimaux sont trouvés.

Entre 0 et 600 ms, les longueurs des plans obtenus par {\it LRTP} et {\it LRTP+J} sont élevées avec une variance élevée parce que 600 ms reste une petite plage de temps de décision pour eux. Inversement, {\it LRTP+I} et {\it LRTP+IJ} ont un petit intervalle (entre 50 et 100 ms) dans lequel la longueur de leur plans solutions baisse. Pour les temps de décision supérieur à 100 ms, la longueur des plans solutions atteint un plateau d'environ 50 actions.

\subsection{Domaine Rovers}

Ce paragraphe présente les résultats obtenus par les quatre algorithmes sur tous les problèmes du domaine Rovers. Il indique le pourcentage de réussite et la qualité de sélection de l'action obtenus dans chaque problème. Deux cas sont considérés: un premier cas où le temps de décision est fixé à 100 ms, et un second cas où le temps de décision est fixé à 500 ms.

\subsubsection{Pourcentage de réussite}

\begin{figure}
\begin{center}
\includegraphics[scale=0.7]{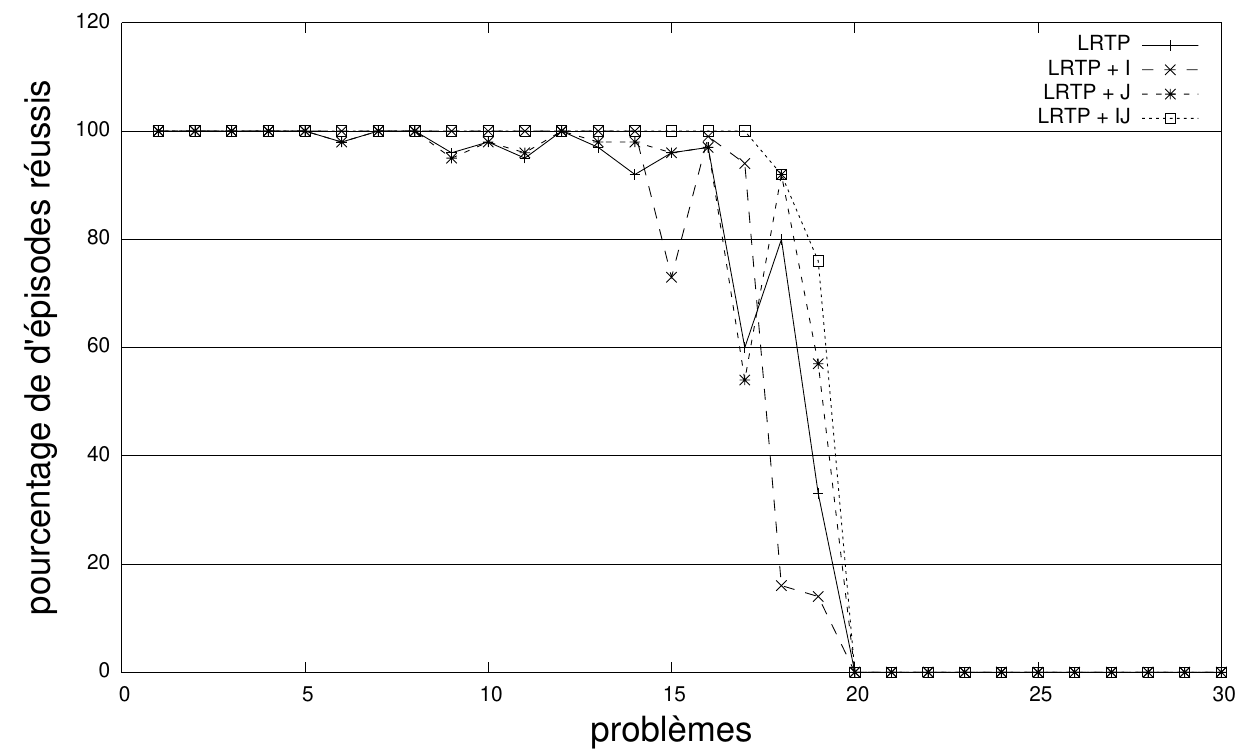}
\caption{Pourcentage de réussite dans le domaine Rovers en fonction du problème lorsque le temps de décision est fixé à 100 ms.}
\label{Fig:success-rovers-100}
\end{center}
\end{figure}

La figure~\ref{Fig:success-rovers-100} montre le pourcentage de réussite de chaque algorithme sur les 40 problèmes du domaine Rovers lorsque le temps de décision est fixé à 100 ms. Nous pouvons voir que tous les algorithmes obtiennent 100\% dans les cinq premiers problèmes. Ensuite, sans amélioration, le pourcentage de réussite reste à plus de 90\% jusqu'au problème 16. Dans les problèmes 17, 18 et 19, il atteint respectivement 60\%, 80\% et 33\% de réussite. Enfin, à partir du problème 20, le planificateur ne trouve pas un plan de solution.

Avec l'amélioration {\it J}, le planificateur obtient des résultats très proches de ceux obtenu sans amélioration. Le pourcentage reste supérieur à 90\% jusqu'au problème 16. Dans les problèmes 17 et 19, il atteint respectivement 54\%, et 57\% de réussite. Dans les problèmes 18, {\it LRTP+J} obtient les mêmes performances que {\it LRTP+IJ} avec 92\% de réussite, ce qui correspond à la meilleure performance des algorithmes de ce problème. Enfin, à partir du problème 20, il ne trouve pas de plan solution.

Avec l'amélioration {\it I}, le planificateur obtient 100\% de réussite jusqu'au problème 14. Il reste supérieur à 90\% jusqu'au problème 17, à l'exception du problème 15 dans lequel le pourcentage est de 79. Pour les problèmes 18 et 19, le pourcentage est inférieur à 20. Enfin, à partir du problème 20, aucun plan solution n'est trouvé.

Les meilleures performances sont obtenues avec les deux améliorations. Nous pouvons voir que 100\% des épisodes se terminent avec un plan solution jusqu'au problème 17. Ensuite, le pourcentage diminue à 92\% dans le problème 18, et 76\% dans le problème 19. Comme tous les algorithmes, il ne trouve pas de plan solution à partir du problème 20.

\begin{figure}
\begin{center}
\includegraphics[scale=0.7]{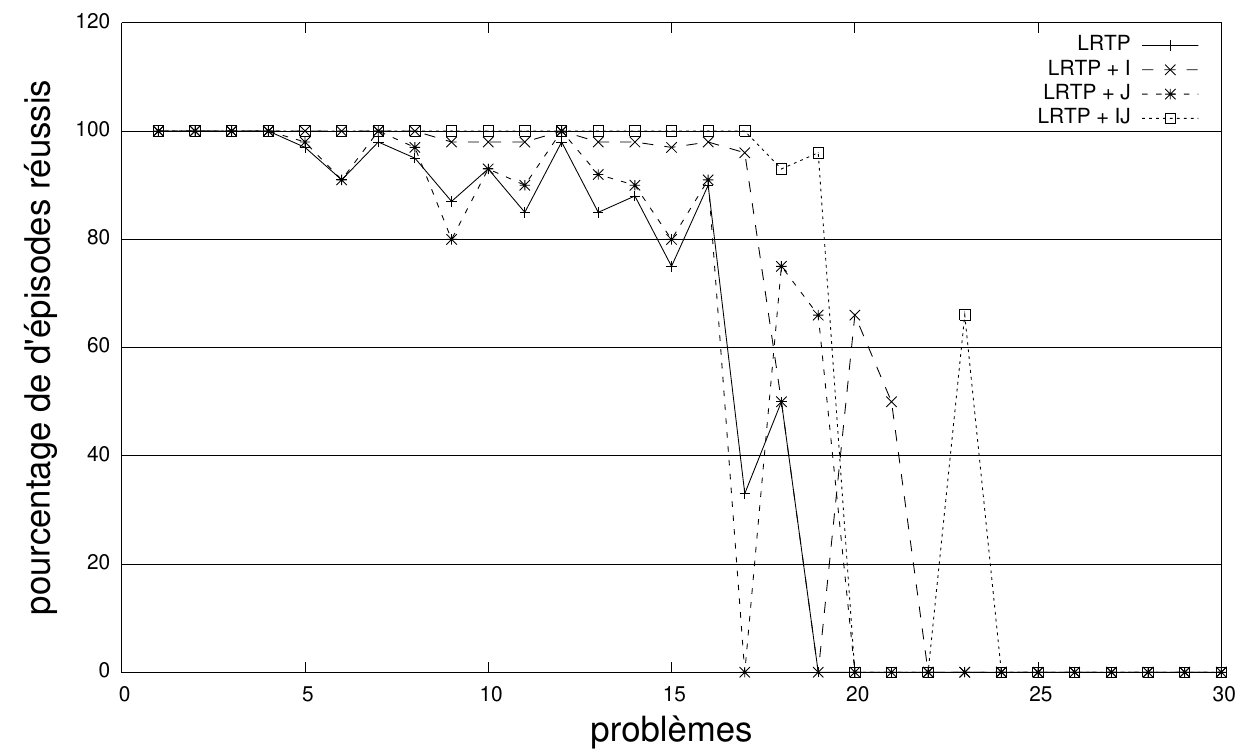}
\caption{Pourcentage de réussite dans le domaine Rovers en fonction du problème lorsque le temps de décision est fixé à 500 ms.}
\label{Fig:success-rovers-500}
\end{center}
\end{figure}

La figure~\ref{Fig:success-rovers-500} montre le pourcentage de réussite de chaque algorithme sur les 40 problèmes du domaine Rovers lorsque le temps de décision est fixé à 500 ms. Nous pouvons voir que tous les algorithmes obtiennent 100\% de réussite dans les quatre premiers problèmes. Ensuite, sans amélioration, le pourcentage de réussite reste à plus de 75\% jusqu'au problème 16. Dans les problèmes 17 et 18, il atteint respectivement 33\% et 50\%. Dans le problème 19, l'algorithme est incapable d'atteindre le but avec 500ms de temps de décision.

Avec l'amélioration {\it J}, le planificateur obtient des performances légèrement meilleures que sans amélioration jusqu'au problème 16, excepté dans le problème 9 où il atteint 80\%. Dans les problèmes 17, 18 et 19, il obtient respectivement 0\%, 75\%, 66\%. A partir du problème 20, le planificateur ne trouve pas de plan solution.

Avec l'amélioration {\it I}, le planificateur obtient plus de 90\% de réussite jusqu'au problème 17. Pour les problèmes au-delà, le pourcentage diminue rapidement. Cet algorithme est le seul qui atteint l'objectif dans les problèmes 20 et 21, avec respectivement 66\% et 50\% de réussite. Dans le problème 22, aucun plan solution n'est trouvé.

Les meilleures performances sont globalement obtenues avec les deux améliorations. Nous pouvons voir que 100\% des épisodes se terminent avec un plan solution jusqu'au problème 17, et 90\% jusqu'au problème 19. A partir du problème 20, il ne trouve pas de plan solution excepté dans le problème 23, où il est le seul algorithme qui atteint le but. Dans ce problème, son pourcentage de réussite est alors de 66\%.

\subsubsection{Qualité de la sélection d'actions}

\begin{figure}
\begin{center}
\includegraphics[scale=0.7]{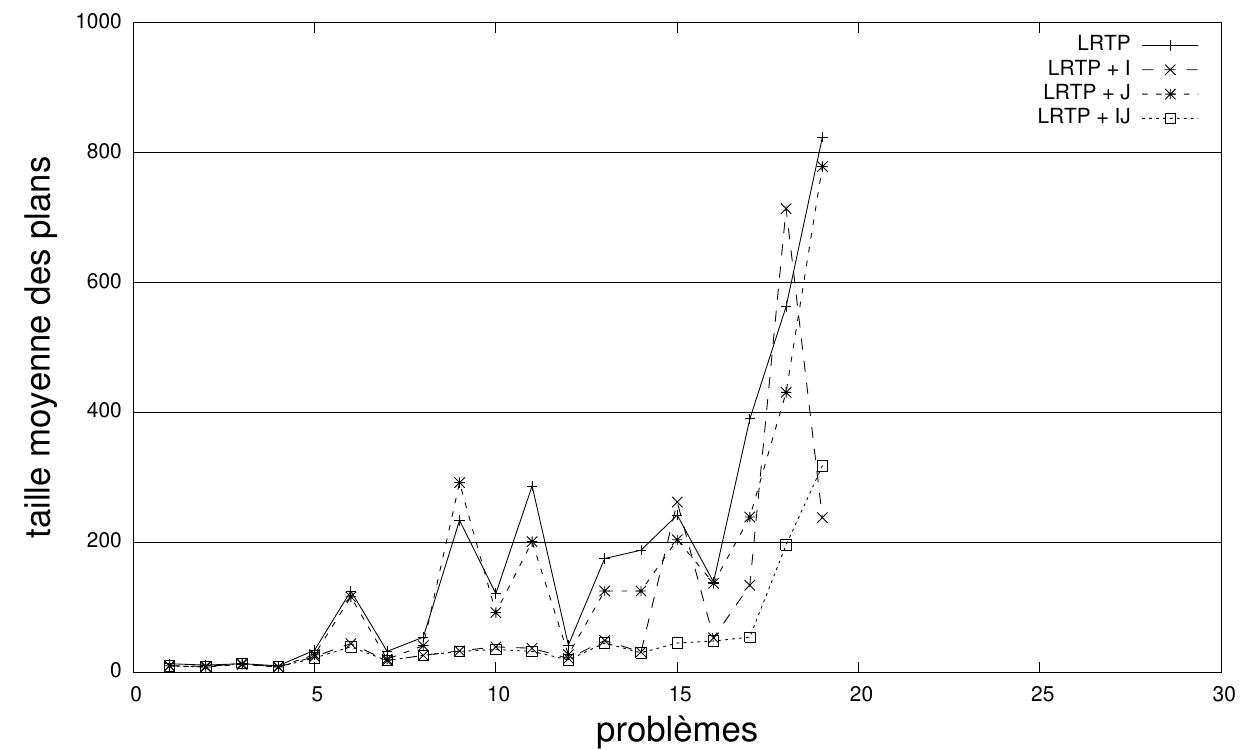}
\caption{Longueur moyenne des plans trouvés dans le domaine Rovers en fonction du problème lorsque le temps de décision est fixé à 100 ms.}
\label{Fig:rovers-AvgPlanLength-100}
\end{center}
\end{figure}

La figure~\ref{Fig:rovers-AvgPlanLength-100} montre la taille moyenne des plans solutions obtenus par les quatre algorithmes dans tous les problèmes du domaine Rovers lorsque le temps de décision est fixé à 100 ms.

Jusqu'au problème 14, et dans le problème 16, les plus petits plans sont trouvés par {\it LRTP+I} et {\it LRTP+IJ} avec des performances très proches. Parallèlement {\it LRTP} et {\it LRTP+J} obtiennent des performances similaires avec une variance plus élevée. Dans les problèmes 15 et 17, les longueurs moyennes des plans trouvés par les algorithmes sont supérieures à 100 excepté pour {\it LRTP+IJ} qui obtient respectivement 45 et 54 actions. Dans les problèmes 18 et 19, les moyennes des longueurs des plans fournis par tous les algorithmes sont élevées. Dans le problème 18, {\it LRTP+IJ} obtient les meilleures performances avec une longueur moyenne de 196 tandis que les autres algorithmes ont des longueurs moyennes supérieures à 400. Dans le problème 19, {\it LRTP+I} obtient les meilleures performances avec des plans contenant 238 actions en moyenne, suivi par {\it LRTP+IJ} avec les plans des 318 actions. Les deux derniers algorithmes fournissent des plans solutions dont la longueur moyenne est supérieure à 800 actions.

\begin{figure}
\begin{center}
\includegraphics[scale=0.7]{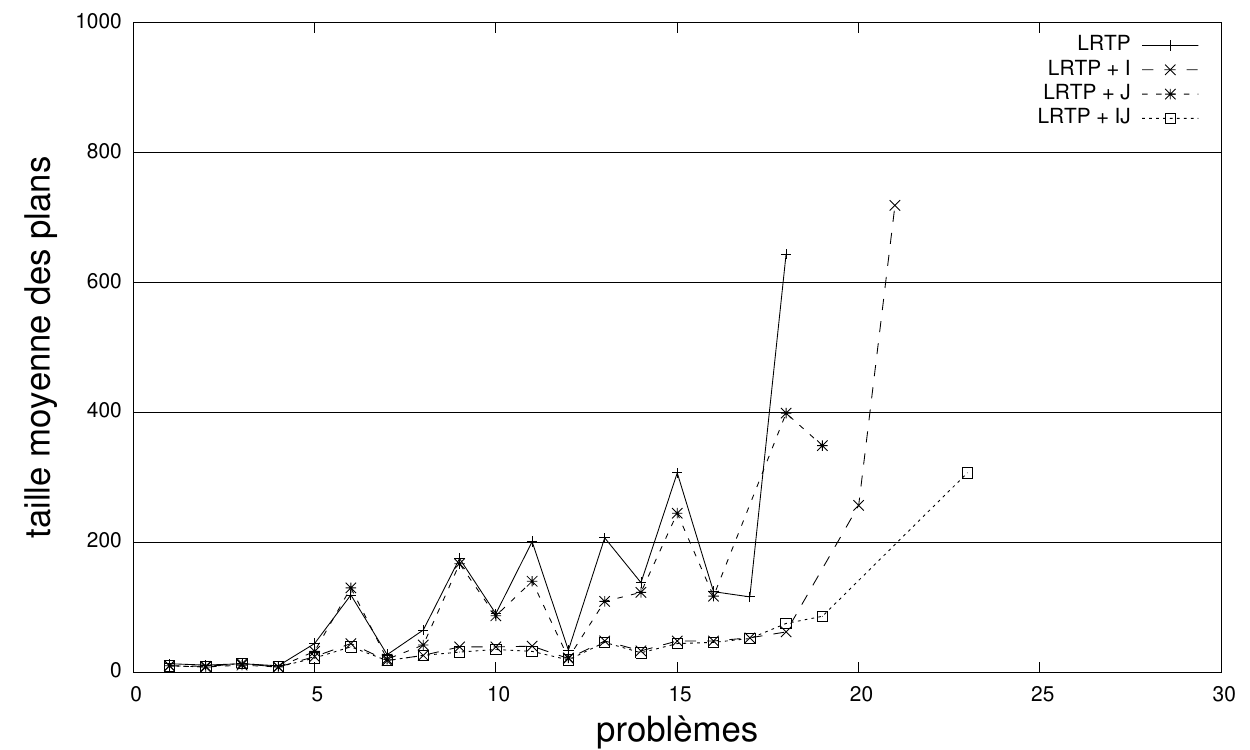}
\caption{Longueur moyenne des plans trouvés dans le domaine Rovers en fonction du problème lorsque le temps de décision est fixé à 500 ms.}
\label{Fig:rovers-AvgPlanLength-500}
\end{center}
\end{figure}

La figure~\ref{Fig:rovers-AvgPlanLength-500} montre la longueur moyenne des plans solutions obtenus par les quatre algorithmes dans le domaine Rovers lorsque le temps de décision est fixé à 500 ms.

Jusqu'au problème 18, les plus petits plans sont trouvés par {\it LRTP+I} et {\it LRTP+IJ} avec des performances très proches. Comme dans la figure précédente, {\it LRTP} et {\it LRTP+J} ont des performances similaires avec une forte variance. Dans le problème 19, seuls {\it LRTP+J} et {\it LRTP+IJ} atteignent l'objectif, avec des longueurs moyennes de plans de respectivement 86 et 349 actions. Seul {\it LRTP+I} atteint l'objectif des problèmes 20 et 21. Seul {\it LRTP+IJ} atteint l'objectif dans le problème 23.

\subsection{Résultats dans différents domaines}

Cette section présente les résultats obtenus par les quatre algorithmes sur tous les problèmes de tous le domaines considérés. Nous avons fixé le temps de décision à une valeur spécifique et avons exécuté les algorithmes dans tous les problèmes de 15 domaines issus de IPC-2, IPC-3, IPC-4 et IPC-5. Ensuite, pour chaque algorithme dans chaque domaine, les mesures des performances fournies sont des moyennes des performances obtenues dans les problèmes de chaque domaine. Chaque expérience contient 100 épisodes. Les épisodes se terminent si l'algorithme atteint l'objectif ou si 500 actions ont été exécutées. Si un algorithme ne peut jamais atteindre le but lors d'une expérience, la mesure de longueur de plan est alors fixée à 500.

Le temps de décision utilisé dépend du domaine considéré. Il est défini suffisamment élevé pour permettre à l'algorithme de faire face à la majorité des problèmes de chacun des domaines. Ainsi, il a été fixé à 100~ms dans le domaine DriverLog, 200~ms dans Pathways, 500~ms dans les domaines Blocksworld, Depots, Elevator, Freecell, Logistics, Pipesworld-no-Tankage, Rovers, Satellite et Zenotravel, et finallement à 1000~ms dans Airport, Opentrack, PSR et TPP.

\subsubsection{Pourcentage de réussite}

\begin{figure}
\begin{center}
\includegraphics[scale=0.87]{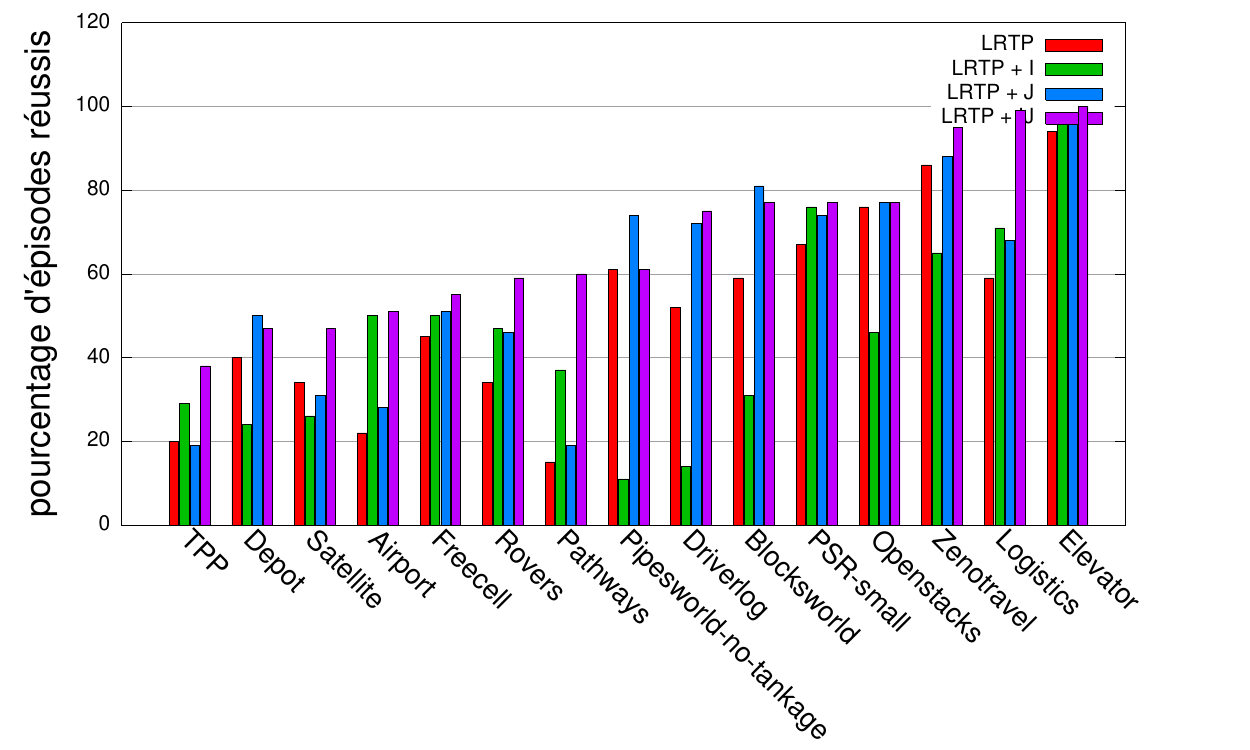}
\caption{Pourcentage de réussite en fonction du domaine.}
\label{Fig:success}
\end{center}
\end{figure}

La figure~\ref{Fig:success} montre le pourcentage de réussite moyen des algorithmes pour chaque domaine étudié. Nous pouvons voir que, dans la majorité des domaines, les meilleures performances sont obtenues avec les deux améliorations. Certaines exceptions sont observées dans Depots, Pipesworld et Blocksworld lorsque l'amélioration {\it J} est utilisée seule, celle-ci donnant des performances légèrement meilleures que lorsque les deux améliorations sont utilisées. Il peut également être observé qu'utiliser l'amélioration {\it J} améliore presque toujours les performances, avec ou sans l'amélioration {\it I}. Une exception est observée dans Satellite. Dans ce cas précis, utiliser une des améliorations diminue le performances alors qu'utiliser le deux permet d'obtenir les meilleures performances.

\subsubsection{Qualité de la sélection d'actions}

\begin{figure}
\begin{center}
\includegraphics[scale=0.87]{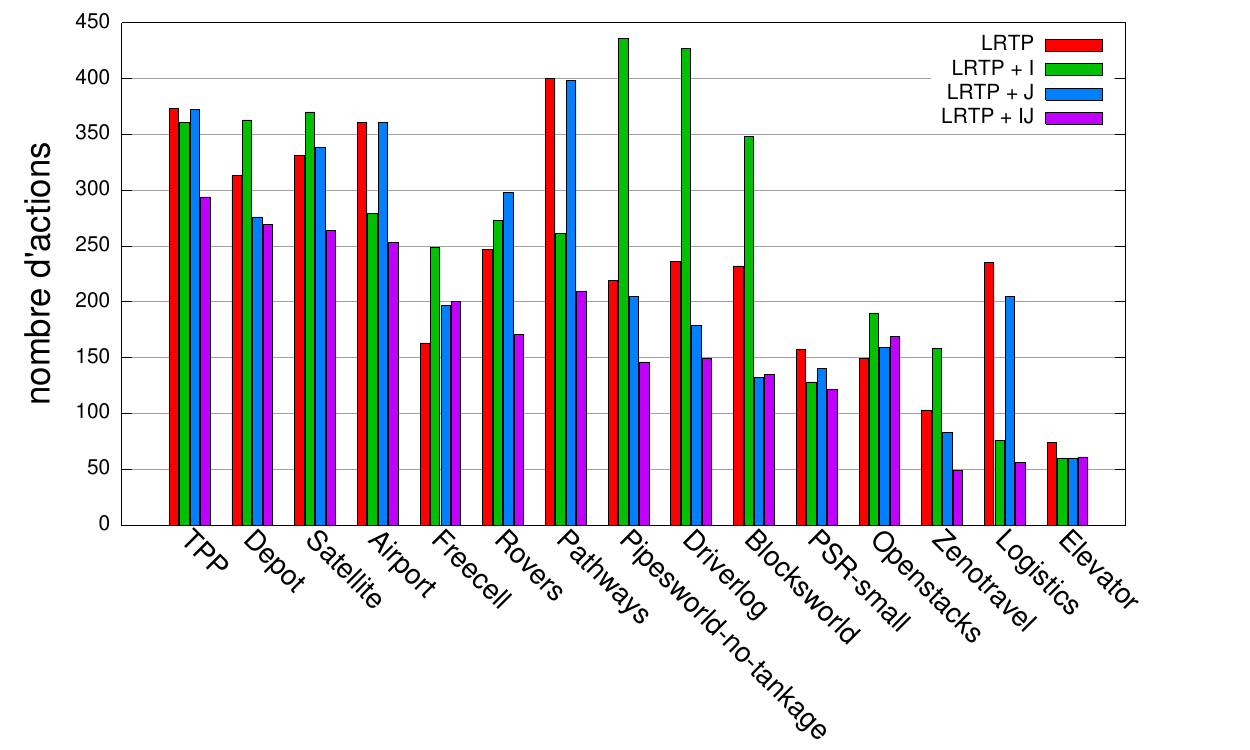}
\caption{Longueur moyenne des plans trouvés en fonction du domaine.}
\label{Fig:all_first_plan_length}
\end{center}
\end{figure}

La figure~\ref{Fig:all_first_plan_length} montre la longueur moyenne des plans solutions trouvés en fonction du domaine. Nous pouvons voir que {\it LRTP+IJ} présente le meilleur comportement. {\it LRTP + I} obtient des résultats médiocres sur Pipesworld, DriverLog et Blocksworld.

\section{Discussion}
\label{sec:discussion}

Tout d'abord, nous souhaiterions commenter les observations globales données par la figure~\ref{Fig:success}. En effet, {\it LRTP+I} n'apparaêt pas clairement comme une amélioration en termes de pourcentage de succès. Ceci peut s'expliquer par le fait que {\it LRTP} adopte une vue globale du problème à résoudre. En revanche, {\it LRTP+I} suit l'ordre de résolution des buts défini par son agenda et possède une vue très locale de ce qu'il fait. On constate que les domaines dans lesquels {\it LRTP+I} est plus mauvais que {\it LRTP}, sont les domaines où l'agenda de buts est très peu informatif.

Par opposition {\it LRTP+J} est clairement une amélioration de {\it LRTP} en termes de pourcentage de succès. Avec de petits temps de décision (suffisamment petit pour que {\it LRTP} soit capables de trouver une solution), agir rapidement apparaît être une bonne stratégie. C'est le cas de {\it LRTP+J} puisqu'il s'engage à exécuter une séquence d'actions et non une seule action comme {\it LRTP}. La séquence d'actions peut bien évidemment être de bonne ou de mauvaise qualité du moment qu'elle est calculée rapidement. Généralement, la séquence est très éloignée de l'optimale. En revanche, le fait de s'engager à exécuter plusieurs actions au cours d'un même temps de décision a pour effet d'accorder plus de temps pour les prochaines décisions et par conséquent leur permettre d'être de meilleure qualité.

Notons également que {\it LRTP+IJ} est clairement une amélioration de {\it LRTP+J}, de {\it LRTP+I} et de {\it LRTP}. Étant donné que les temps de décision sont courts et qu'il est intéressant d'agir rapidement, les décisions résultant de la recherche dirigée par l'agenda de buts sont de meilleure qualité que lorsque la recherche essaie d'embrasser tout le problème de manière globale. Ceci montre qu'en activant les sauts, il est mieux de s'engager sur une séquence d'actions. La figure \ref{Fig:all_first_plan_length} illustre que {\it LRTP+IJ} produit des plans solutions plus courts que les plans produit par {\it LRTP+J}. Par conséquent, {\it LRTP+IJ} génère des plans plus courts et plus rapidement tout en obtenant de meilleurs pourcentages de succès. Nous pouvons en conclure que {\it I} et {\it J} doivent être utilisés simultanément.

Pour conclure, soulignons que {\it LRTP+J} est différent de l'algorithme {\it LRTS} avec sauts proposé par \cite{bulitko:06}. Lorsqu'une séquence d'actions est trouvée, {\it LRTS} fait l'hypothèse que la séquence d'actions sera exécutée dans le même temps de décision. Cette supposition n'est pas très réaliste dans la mesure où l'exécution d'une action peut ne pas être instantanée. Ceci a pour effet que {\it LRTS} ne tire pas parti de ses sauts. Inversement, planifier une séquence d'actions comme le fait {\it LRTP} en exécutant les actions une à une permet de capitaliser du temps pour prendre de meilleures décisions, trouver des plans solutions plus courts et indirectement gagner du temps.

\section{Conclusion}
\label{sec:conclusion}

Dans cet article, nous avons présenté un modèle pour la planification temps réel, appelé {\it LRTP}, nous permettant de proposer deux améliorations pour la phase de sélection d'actions: la première amélioration {\it I} s'appuie sur un agenda de buts pour classer les buts par ordre de difficulté croissante et tente ensuite de les résoudre de manière incrémentale; la seconde amélioration {\it J} consiste à choisir à chaque pas de décision une séquence d'actions qui sera exécutée et non plus une seule. L'amélioration {\it I} accélère le processus de sélection d'actions mais diminue le pourcentage de succès. L'amélioration {\it J} ne modifie ni le pourcentage de succès, ni la qualité des plans solutions trouvés. La contribution majeure de notre travail est de montrer qu'utilisées simultanément ces deux améliorations permettent à {\it LRTP} d'obtenir des résultats qui surpassent les autres versions de {\it LRTP}. Le pourcentage de succès est meilleur, les plans solutions plus courts, et par conséquent le temps nécessaire à leur exécution. En outre, ces deux améliorations combinées permettent à {\it LRTP} de travailler avec des temps de décisions très bas (de l'ordre de quelque centaines de millisecondes pour {\it LRTP} sans amélioration à quelques millisecondes pour pour {\it LRTP+IJ}). Pour autant que nous le sachions, c'est la première fois que l'agenda de buts est utilisé avec succès dans le cadre de la planification temps réel.

Plusieurs pistes intéressantes de recherche sont ouvertes par ce travail. Tout d'abord, il serait intéressant d'étudier précisément l'apprentissage en examinant d'autres mécanismes d'apprentissage pour la fonction heuristique et en mettant en place un mécanisme d'apprentissage de l'agenda de buts. De plus, nous n'avons considéré dans nos tests qu'une seule fonction heuristique. Finalement, il serait aussi intéressant de proposer et de comparer d'autres mécanismes de sélection d'actions s'appuyant par exemple sur la technique de <<~landmarks planning~>> proposée par \cite{richter:08a}, ou encore sur des techniques de stratification telle que \cite{chen:09}.

\end{document}